\documentclass[conference,10pt]{IEEEtran}
\IEEEoverridecommandlockouts
\usepackage[utf8]{inputenc}
\usepackage{setspace}

\usepackage[acronym]{glossaries}
\usepackage{subcaption}  
\usepackage{multirow}
\usepackage{makecell} 

\usepackage[linesnumbered,ruled,vlined]{algorithm2e}
\newcommand{\nosemic}{\renewcommand{\@endalgocfline}{\relax}}
\newcommand{\dosemic}{\renewcommand{\@endalgocfline}{\algocf@endline}}
\let\oldnl\nl
\newcommand{\nonl}{\renewcommand{\nl}{\let\nl\oldnl}}
\usepackage{bbm}

\usepackage{tikz}
\usepackage{pgfplots}
\usepgfplotslibrary{groupplots}
\pgfplotsset{compat=newest}
\usepgfplotslibrary{colorbrewer, groupplots, patchplots}
\pgfplotsset{
    compat=newest,
    grid=both,
    legend style={
        font=\scriptsize,
        fill opacity=0.7,
        draw opacity=1,
        text opacity=1,
        draw=white!15!black
    },
    legend columns=3,
    width=0.77\columnwidth,
    scale only axis,
    height=2cm,
    yminorticks=false,
    xminorticks=false,
    label style={font=\scriptsize},
    title style={font=\scriptsize},
    tick align=outside,
    tick pos=left,
    tick style={color=black},
    tick label style={font=\scriptsize},
    grid style={line width=.1pt, draw=gray!20},
    major grid style={line width=.1pt, draw=gray!20},
    plot coordinates/math parser=false,
    xlabel style={
        at={(axis description cs:0.5,-0.22)},  
        anchor=north
    },
    ylabel style={
        at={(axis description cs:-0.21,0.5)},  
        anchor=south,
        align=center,
    },
    title style={
        at={(axis description cs:0.5,0.9)},  
        anchor=south,                         
        font={\scriptsize}
    }
}
\usetikzlibrary{shapes,positioning}
\usepackage{amsmath}
\usepackage{amsthm}
\usepackage{amssymb}
\usepackage{amsfonts}
\usepackage{dsfont}
\usepackage{balance}
\usepackage[subnum]{cases}
\usepackage[nocomma]{optidef}
\usepackage[normalem]{ulem}
\usepackage{mathtools}
\usepackage{url}

\usepackage{enumitem}
\usepackage[acronym]{glossaries}
\newacronym{ae}{AE}{autoencoder}
\newacronym{ap}{AP}{access point}
\newacronym{ad}{AD}{anomaly detection}

\newacronym{cl}{CL}{continual Learning}

\newacronym{dnn}{DNN}{Deep Neural Network}
\newacronym{en}{EN}{easy negative}
\newacronym{es}{ES}{edge server}
\newacronym{fd}{FD}{fault detection}
\newacronym{fp}{FP}{false positive}
\newacronym{fpr}{FPR}{false positive rate}
\newacronym{fn}{FN}{False Negative}

\newacronym{hn}{HN}{hard negative}

\newacronym{iot}{IoT}{Internet of Things}
\newacronym{iid}{i.i.d.}{independent and identically distributed}

\newacronym{kpi}{KPI}{key performance indicator}
\newacronym{mcu}{MCU}{micro controller unit}
\newacronym{ml}{ML}{machine learning}

\newacronym{ocl}{OCL}{Online Continual Learning}
\newacronym{ocsvm}{OCSVM}{One-Class SVM}
\newacronym{tinyml}{TinyML}{Tiny Machine Learning}
\newacronym{tp}{TP}{true positive}
\newacronym{tpr}{TPR}{true positive rate}

\newacronym{tdr}{TDR}{true (false) discovery rate}

\DeclareMathAlphabet{\mathcal}{OMS}{cmsy}{m}{n}
\SetMathAlphabet{\mathcal}{bold}{OMS}{cmsy}{b}{n}

\definecolor{darkgreen}{rgb}{0.0,0.5,0.0}

\allowdisplaybreaks

\title{Online Continual Learning for Anomaly Detection in IoT under Data Distribution Shifts}
\author{\IEEEauthorblockN{Matea Marinova\IEEEauthorrefmark{1}, Shashi Raj Pandey\IEEEauthorrefmark{2}, Junya Shiraishi\IEEEauthorrefmark{2}, Martin Voigt Vejling\IEEEauthorrefmark{2}, Valentin Rakovic\IEEEauthorrefmark{1}, and Petar Popovski\IEEEauthorrefmark{2}\IEEEauthorrefmark{1}}
\thanks{The work was supported by DFF-Forskningsprojekt1 ``NETML” (nr. 4286-00278B), MSCA-PF ``NEUTRINAI" (nr.~101151067), and Villum Investigator Grant ``WATER" (nr. 37793).}
\IEEEauthorblockA{\IEEEauthorrefmark{1}Faculty of Electrical Engineering and Information Technologies, Ss. Cyril and Methodius University in Skopje,\\ 1000 Skopje, North Macedonia (\{mateam, valentin\}@feit.ukim.edu.mk)}
\IEEEauthorblockA{\IEEEauthorrefmark{2}Dept. of Electronic Systems, Aalborg University, Aalborg, Denmark (\{srp, jush, mvv, petarp\}@es.aau.dk)}
}
\begin{document}
\maketitle

\begin{abstract}
In this work, we present OCLADS, a novel communication framework with continual learning (CL) for Internet of Things (IoT) anomaly detection (AD) when operating in non-stationary environments. As the statistical properties of the observed data change with time, the on-device inference model becomes obsolete, which necessitates strategic model updating. OCLADS keeps track of data distribution shifts to timely update the on-device IoT AD model. To do so, OCLADS introduces two mechanisms during the interaction between the resource-constrained IoT device and an edge server (ES): \emph{i}) an intelligent sample selection mechanism at the device for data transmission, and \emph{ii}) a distribution-shift detection mechanism at the ES for model updating. Experimental results with TinyML demonstrate that our proposed framework achieves high inference accuracy while realizing a significantly smaller number of model updates compared to the baseline schemes.
\end{abstract}

\section{Introduction}
\Gls{iot} \gls{ad} is critical for a variety of practical applications, including status monitoring of industrial machinery \cite{chevtchenko2023anomaly} and health monitoring via wearable devices \cite{mathivanan2024comprehensive}. Specifically, it enables identification of abnormal behavior in individual devices or entire \gls{iot} systems. \Gls{ml}-based approaches have significantly improved the efficiency of \gls{iot} \gls{ad}. However, limited labeled data and computational resources remain key bottlenecks for practical deployment \cite{chevtchenko2023anomaly}. 
 
Lightweight \gls{ml} models, such as \gls{tinyml}, to some degree, address the challenges of constrained computation, allowing estimates of an anomalous event based on complex high dimensional input data or data sequences observed by each \gls{iot} device~\cite{banbury2020benchmarking}. The problems of constrained computation and limited labeled data can be addressed by an orchestrated approach~\cite{frederiksen2025link}: the \gls{es} collects centrally large samples from the connected devices, trains a better \gls{ad} model and subsequently shares it with the \gls{iot} devices. However, in non-stationary, dynamic environments, the traditional approach of using static ML models for \gls{ad} performs poorly due to lack of adaptability to changing input data distributions.

By continually adapting the available model to the current data distribution, and retaining the knowledge of previous ones, \Gls{cl} methods aim to address this challenge \cite{ravaglia2021tinyml,oclmetrics}. Therein, low-power \gls{iot} devices can leverage \gls{cl} to adapt to changing data distributions, e.g., \cite{ravaglia2021tinyml} proposes a novel quantized latent replay-based \gls{cl} method, while \cite{paissan2024structured} reduces the memory and computational costs of backpropagation in latent replay-based \gls{cl} using a sparse update strategy. Similarly, in~\cite{pandey2024onboard}, genetic algorithms integrated with \gls{tinyml} were used to enable class-incremental learning on tiny devices. However, these strategies consider \gls{cl} directly on resource-constrained devices, adding an additional computational load with frequent model updates. In contrast to these approaches, we aim to develop a collaborative framework for online \gls{cl} targeting \gls{ad} tasks under non-stationary conditions and addressing communication efficiency, i.e., deciding \textit{when} and \textit{how} to update the on-device \gls{iot} \gls{ad} model. The most related work introduced an event-driven \gls{cl} framework in \gls{iot} networks~\cite{frederiksen2025link}, but it does not address update scheduling while considering the impact of the distribution shifts to the \gls{ad} task. 
 
In this work, we propose a framework for \textit{Online \gls{cl} for \gls{ad} under data distribution Shifts} (OCLADS) in distributed settings. Our proposed framework integrates an intelligent sample selection mechanism along with an algorithm for distribution shift detection based on a hypothesis testing framework. As such, the \gls{es} shares a new model with the \gls{iot} device only when a shift occurs and the model update yields a substantial performance gain. This results in a more efficient use of the wireless medium by avoiding unnecessary communication overhead with arbitrary model updates policy. To the best of our knowledge, this is the first work that integrates distributed \gls{tinyml} and online \gls{cl}-based \gls{ad} under non-stationary conditions, along with communication-efficient data shift-aware model updates and importance-driven sample selection.

\section{System Model} \label{sec:systemmodel}
We consider an online on-device \gls{ad} setup, where a single \gls{iot} device conducts \gls{ad} and communicates with the \gls{es} through the wireless medium. The \gls{iot} device consists of a \gls{mcu}, a radio interface, and vision sensors. The \gls{iot} device installs a \gls{tinyml} model to continuously perform on-device inference against the upcoming data batch.
In order to realize the deployment of \gls{tinyml} in practical non-stationary environment, this paper applies a \gls{cl} setup considered in~\cite{frederiksen2025link,hawk}. In this framework, the deployed model is updated continuously by sharing/receiving data/model with/from the \gls{es}, where, due to the memory constraint at the \gls{iot} device, the updating of the model is done at the \gls{es} using data collected by the \gls{iot} device.

\subsection{Observations Model with Distribution Shift}
The time is divided into communication rounds, indexed by $i \in \mathcal{N}=\{1, 2, \ldots, N\}$.
In each communication round $i$, the \gls{iot} device is able to access a batch, $\mathcal{X}_i=\{{\bf{x}}_i^j\}_{j=1}^{B_i}$, of $B_{i}$ incoming data samples where ${\bf{x}}_i^j \in \mathbb{R}^{M}$ is $M$-dimensional data such as an image. We assume samples are \gls{iid} within each batch $i$, thereby following distribution $P_i$, as
\begin{equation}
 {\bf{x}}_i^j \overset{\mathrm{{i.i.d}}}{\sim} P_i, \forall j \in \mathcal{B}_i,
\end{equation}
where $\mathcal{B}_i = \{1, 2, \ldots, B_i\}$.

\subsubsection{Anomaly Process and its Detection}

For each batch $i$, we model anomaly presence as an unknown binary sequence:
\begin{equation}
 y_i^j = \begin{cases}0,& \text{No anomaly},\\1,& \text{Anomaly}, \end{cases}
\end{equation}
where $y_i^j, i \in \mathcal{N}, j \in \mathcal{B}_i$ is a ground truth label of sample $j$ at the $i$-th communication round. 
The \gls{tinyml} model installed into the \gls{iot} device at the $i$-th round, denoted as $\mathcal{M}_i: x_i^j \rightarrow \hat{y}_i^j \in \{0, 1\} $, can estimate the corresponding labels as $\hat{y}_i^j, j \in \mathcal{B}_i$ for each sample $x_j^i$ gathered at the $i$-th batch. 
The samples estimated as non-anomalies, i.e., samples with $\hat{y}_i^j = 0$, can be divided into \glspl{en} and \glspl{hn} as in~\cite{hawk}. The \glspl{hn} represent samples that are close to the decision boundary, which makes them both more susceptible to misclassification and more informative than \glspl{en} in terms of improving the model accuracy~\cite{hawk}.

\subsubsection{Distribution Shifts}
This setup considers non-stationary conditions, in which data distribution shifts might happen across consecutive batches, such as label and covariate shifts. In general, they can be either abrupt or gradual~\cite{baby2025adapting}. The former represents an abrupt change in the data distribution between two consecutive rounds, i.e., a shift between time $t-1$ and $t$, while the latter represents a rather slower change in the distribution over time. This work focuses on abrupt covariate shifts\footnote{Note that our framework can be applicable to any type of distribution shift, including label shift}. in the input distribution of the $(i-1)$-th and $i$-th batch. Formally the covariate shift can be described as follows:   \begin{equation}
 P_i ({\bf{x}}) \neq P_{i-1} ({\bf{x}}), \text{with}~P_i(Y|{\bf{x}}) = P_{i-1}(Y|{\bf{x}}),~\forall~i,
\end{equation}
where ${\bf{x}}\in\mathbb{R}^M$ is an $M$-dimensional input and $Y$ is a one-dimensional class output. 
Here, we assume non-stationary data/batch generation processes $\{P_i\}_{i \geq 0}$ and that the evolution of $P_i$ can occur at any time, unknown to the \gls{iot} device. Furthermore, we assume that the \gls{iot} device cannot detect the distribution shift at the device side due to memory constraints. 

\subsection{Communication Model}
Each communication round can be divided into two parts: an uplink data transmission period, which is followed by a downlink model transmission period. For simplicitiy, we assume that both uplink/downlink transmission are error-free. In each communication period, the \gls{iot} device first decides the subset of samples to be transmitted to the \gls{es} $\mathcal{K}_i \subseteq \mathcal{X}_i$, with its cardinality $K_i = |\mathcal{K}_i |$. After receiving data from the \gls{iot} device, the \gls{es} applies \gls{cl} using the $i$-th received data $\mathcal{R}_i \subseteq \mathcal{K}_i$ and previously received data $\bigcup_{n= 1}^{i-1} \mathcal{R}_n$.
Note that $\mathcal{R}_i = \mathcal{K}_i$ due to the assumption of an error-free channel. Then, the \gls{es} decides whether to transmit a new model update to the \gls{iot} device based on the collected data. If the condition is satisfied, it transmits the updated model to the device during the downlink communication period; otherwise, it suppresses model transmission. This reduces airtime, but the inference accuracy of the \gls{iot} device might be degraded if its model is obsolete. After the downlink transmission period is completed, each node resumes the \gls{ad} task, and a new round begins. 

The timing of the \gls{tinyml} model updates is critical to sustaining high on-device inference accuracy for \gls{ad}, especially under limited device-side energy and channel resources. Intuitively, frequent (infrequent) model updates lead to high (low) inference accuracy as the \gls{iot} device maintains fresh (obsolete) model. Furthermore, fewer updates lead to more efficient channel usage compared to frequent model updates. This paper aims to characterize the trade-off between inference accuracy and the number of \gls{tinyml} model updates, and to demonstrate the advantages of intelligently scheduling model updates. Our framework provides a design guideline for \gls{cl} on resource-constrained \gls{iot} devices. The specific mechanism for uplink/downlink transmission is explained in Sec.~\ref{sec: Proposed_Framework}.

\section{Proposed Framework} \label{sec: Proposed_Framework}
We introduce a framework, called  OCLADS, that can realize online anomaly detection through the exchange of data and model updates with the \gls{es} under distribution shifts. To control the balance between the communication costs and inference accuracy, we introduce two key mechanisms for the \gls{iot} device and the \gls{es}: 1) Informative Sample Selection Mechanism (see Sec.~\ref{sec: batch_selection}) and 2) Intelligent Model Update Method, which relies on a hypothesis testing framework (see Sec.~\ref{sec:Hypothesis_test}).

\begin{figure}[t!]
    \centering
    \includegraphics[width=\linewidth]{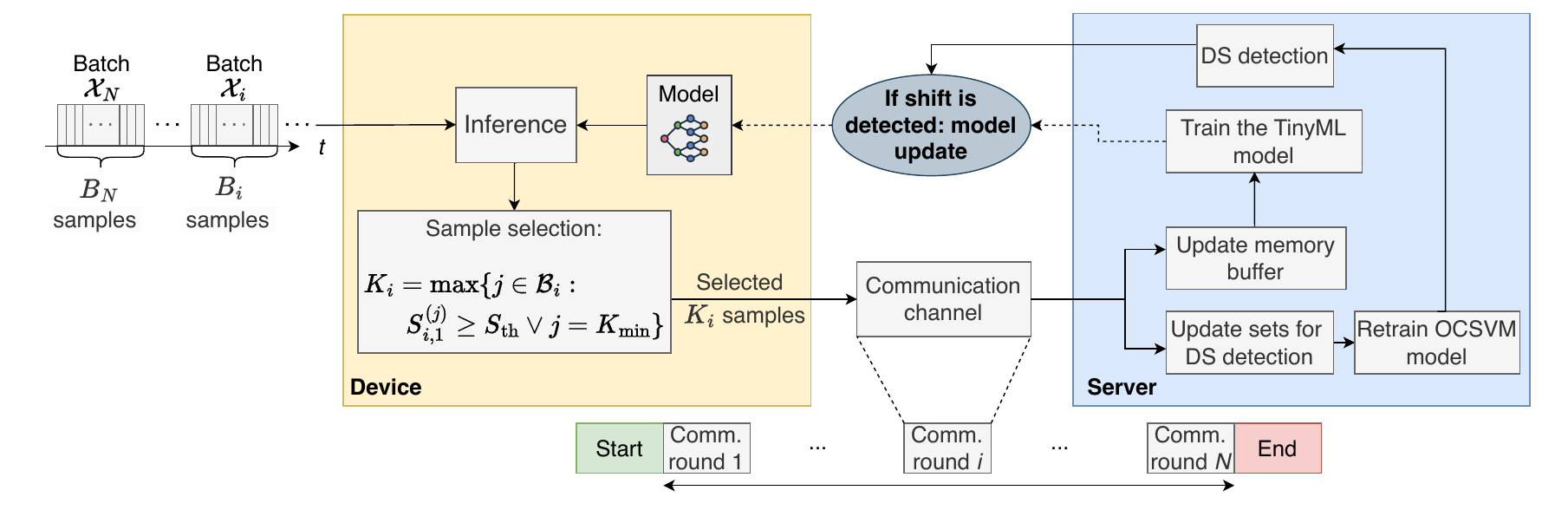}
    \caption{Overview of the OCLADS framework for anomaly detection in online \gls{cl} under non-stationary conditions.}
    \label{fig:framework}
\end{figure}

Fig.~\ref{fig:framework} illustrates the overview of OCLADS, which consists of two interconnected components, each handled by a distinct algorithm: the one for \gls{ad} conducted on the \gls{iot} device (Inference), and the other for detecting data distribution shifts at the \gls{es} side. The \gls{ad} algorithm incorporates the intelligent data selection mechanism, whereas the latter one controls when device-side model updates occur. Notably, the \gls{ad} performance depends on the accuracy of the shift detection method. If a shift is missed, the \gls{iot} device continues to use an obsolete model, which reduces the inference accuracy, and degrades the data selection mechanism. Since the \gls{es} trains the model using the subset of samples selected by the device, poor selection may affect the quality of future updates as well. This creates a feedback loop in which poor shift detection further worsens both \gls{ad} performance and data selection in subsequent iterations.

The proposed OCLADS framework aims to reduce the communication costs without significantly degrading the model performance. It focuses on the most critical operating point, where both phenomena (anomalies and distribution shifts) occur simultaneously.

\subsection{Sample Selection and Uplink Data Transmission}\label{sec: batch_selection}
In each communication round, the \gls{iot} device selects a subset of informative samples for transmission to the \gls{es}. To improve the model performance via \gls{cl} under the low-power requirements of \gls{iot} devices, the applied sample selection mechanism should consider two crucial aspects:   (\emph{a}) the number of transmitted samples and (\emph{b}) each sample's importance for model training. Therefore, it uses an uncertainty-based importance metric that reflects their potential contribution to the model accuracy if used for training. Specifically, since the anomalies represent rare events, the sample selection mechanism should prioritize sending them to the \gls{es}. Furthermore, the selected data should also include \glspl{hn}, which improve the model's ability to distinguish subtle differences, contrary to \glspl{en} that provide little contribution to the training process.

For initial calibration, in the first $L$ communication rounds, the \gls{iot} device turns off the informative sample selection mechanism (described later), and transmits all incoming data to the \gls{es} to ensure that the rare events are included in the training dataset. After finishing these initial communication rounds, i.e., for $i > L$, the \gls{iot} device applies the proposed sample selection method for online \gls{cl} under distribution shifts, as detailed below. 
In order to realize informative data transmission while minimizing wasteful sample transmission, we introduce a transmission threshold $S_{\mathrm{th}}$ to decide whether a sample ${\bf{x}}_i^j$ should be selected or not. Specifically, we leverage the softmax probability obtained from the on-device inference using the \gls{tinyml} model. Let $S_{i,1}^j = g({\bf{x}}_i^j) \in [0,1]$ denote the softmax probability of ${\bf{x}}_i^j$ belonging to the anomalous class, and define ${\bf{x}}_i^{(1)},\dots,{\bf{x}}_i^{(B_i)}$ as the permutation of the data samples obtained by ordering them decreasingly according their anomaly scores, meaning $S_{i,1}^{(1)} \geq \cdots \geq S_{i,1}^{(B_i)}$, with $S_{i,1}^{(j)}=g({\bf{x}}_i^{(j)})$. In the $i$-th round, the \gls{iot} device assigns the subset of selected samples to be transmitted at the $i$-th uplink transmission phase to be $\mathcal{K}_i = \{{\bf{x}}_i^{(j)}\}_{j=1}^{K_i}$ where $K_i = {\rm max}\{j \in \mathcal{B}_i : S_{i,1}^{(j)} \geq S_{\rm th} \lor j = K_{\rm min}\}$. In this way, priority is given to transmitting data samples with the highest anomaly scores. It is ensured that all samples with anomaly scores above the threshold $S_{\rm th}$ are transmitted, and that at least $K_{\rm min}$ samples are transmitted in each round.

The server applies a replay-based \gls{cl} method, specifically the Experience Replay baseline from~\cite{iblurry}, to train its model using the received samples along with previously stored ones in the buffer. We assume that accurate labels become available to the server in time to perform the training, e.g., by human labeling or by observing an independent process revealing the labels post-hoc.

\subsection{Hypothesis Testing Framework for Model Updates}\label{sec:Hypothesis_test}
In this section, we present the hypothesis testing framework, assuming that the \gls{es} has received data $\mathcal{R}_i= \{\hat{\bf{x}}_{i}^{j}\}_{j=1}^{|\mathcal{R}_i|}$ in the $i$-th communication round. The null hypothesis for batch $i$ can be defined as
\begin{equation}
    H_0^i: \hat{\bf{x}}_{i}^{j} \sim \hat{P}_{i-1}, \forall~j\in \mathcal{R}_i. \label{null}
\end{equation}
where $\hat{P}_{i-1}$ denotes the data distribution of the data received in the previous batch $i-1$. That is, for simplicity, we focus only on detecting distribution shifts between two consecutive batches. Note that $\hat{P}_{i-1}$ corresponds to $P_{i-1}$ after applying the device-side selection rule based on the softmax thresholding and the top-$K_\mathrm{min}$ fallback. The null hypothesis means that there is no distribution shift. As such, if the hypothesis is rejected, the \gls{es} detects the distribution shift, and triggers sending the model update to the \gls{iot} device. Obviously, the choice of $K_{\text{min}}$ and $S_{\rm th}$ is critical for the accuracy of the algorithm, i.e., lower (higher) $K_i$ values will result in less (more) accurate shift detection with lower (higher) communication costs.

To test hypothesis $H_0^i$ at a specified significance level $\alpha \in (0, 1)$ without making distributional assumptions, we employ a two-sample testing technique as presented in \cite{vejling2025conformal} due to its flexibility, low complexity, and high performance. The method proceeds as follows: (i) A one-class classifier is trained on data from previous batches stored in the buffer at the \gls{es} side (excluding $\mathcal{R}_{i-1}$), specifically we use the \gls{ocsvm} method, resulting in a score function $\hat{s}(\cdot)$; (ii) The scores $\hat{s}_i^j = \hat{s}(\hat{\bf{x}}_i^j)$ are computed on the calibration data, $\mathcal{R}_{i-1}$, and test data, $\mathcal{R}_{i}$; (iii) For $i'\in \{i-1, i\}$, empirical cumulative distribution functions (CDFs) are computed as
$\hat{F}_{i'}(x) = \sum_{j=1}^{|\mathcal{R}_{i'}|} \mathbbm{1}(\hat{s}_{i'}^{j} \leq x)/|\mathcal{R}_{i'}|$
where $\mathbbm{1}(\cdot)$ is the indicator function, and a test statistic
$T_{L_2} = \int_{-\infty}^\infty |\hat{F}_{i-1}(x) - \hat{F}_{i}(x)|^2 {\rm d}x$,
is computed; (iv) following a permutation testing approach,
$I$ permutations are created by randomly assigning data samples from $\mathcal{R}_{i-1}$ and $\mathcal{R}_{i}$ to new splits $\mathcal{R}_{i-1}^{(k)}$ and $\mathcal{R}_{i}^{(k)}$ for $k=1,...,I$, and statistics $T_{L_2}^{(k)}$ are evaluated; (v) Finally, a p-value is computed as
\begin{equation}
    p_i = \frac{1+\sum_{k=1}^{I} \mathbbm{1}(T_{L_2}^{(k)} \geq T_{L_2})}{I+1}.
\end{equation}
The resulting p-value is then compared with the chosen significance level $\alpha$. If $p_i \leq \alpha$, a data distribution shift is detected and $H_0^{i}$ is rejected. This indicates that choosing a higher $\alpha$ would result in more shift detections and higher channel use. Conversely, using a lower value can reduce channel use, i.e., the number of device-side updates, but may lead to more undetected shifts.

Alternative models to \gls{ocsvm} can be employed for computing scores, and other test statistics than the $L_2$-norm may also be explored \cite{vejling2025conformal}. Additionally, other techniques such as the maximum mean discrepancy \cite{Gretton2012:Kernel} could also be used. However, numerical experiments validate that the implemented distribution shift test performs sufficiently well in the contexts considered in this paper.

\section{Experimental Setup}
This section describes the experiments, including the datasets and models used, the method used to simulate distribution shifts, the baseline schemes, and the numerical results.

\subsection{Datasets and Models}

The experiments are performed using the CIFAR-10 \cite{cifar10} and SVHN \cite{svhn} datasets. Both are modified for \gls{ad}, where the anomalous class in the datasets represents 7\% of the entire data. For CIFAR-10, the class \textit{`cat'} is treated as the anomalous class, whereas all the remaining classes are grouped to form the normal class. Therefore, the resulting set consists of 48280 samples in total, out of which 3380 samples are anomalous. Similarly, the SVHN dataset is modified by treating digit `1' as anomalous, while the rest of the digits represent the normal class. This modified SVHN variant includes 63757 samples with 4463 of them being anomalies.

In this setup, we employed the MCUNet model \cite{mcunet}, which is considered suitable for resource-constrained devices. Specifically, the experiments are conducted using the MCUNet-in1 pretrained model. The images from both datasets were first resized to dimensions $96\times96\times3$ as the input resolution originally specified for this architecture. Model quantization was not applied, since our findings remain valid regardless, and the MCUNet's quantization performance has already been extensively studied.

\subsection{Simulation Model and Baseline Schemes} \label{sdds}

To simulate data distribution shifts we use image corruption functions from \cite{imagecorruption}. Specifically, we apply Gaussian noise, fog, frost, and brightness as corruption types with severity levels 0 (no corruption), 3 and 5 (maximum severity). All samples from a given batch have the same corruption type and level. This ensures that distribution shifts occur only between batches, but not between samples within the same batch. A candidate shift is generated with probability 15\% per batch, which corresponds to a change in the severity level and/or corruption type. However, the effective number of shifts is lower since the candidate shifts must satisfy a minimum inter-shift interval of 5 batches. Additionally, transitions from severity level 0 to 0 are ignored. Note  that the stated probability serves as a scenario specific parameter that can be adjusted. Once a shift occurs, the selected corruption and severity are maintained in subsequent batches until another shift occurs.

The online training is conducted on the \gls{es} using a replay-based method. Prior to initiating \gls{cl}, we fine-tune the pretrained MCUNet model on a small dataset of 120 samples (100 normal and 20 anomalous). This step, inspired by the bootstrap strategy in \cite{hawk}, aims to provide a better initial representation for both classes. The resulting model parameters are then used to initialize the neural network for the subsequent \gls{cl} phase, thereby improving early-stage \gls{ad} and stability under distribution shifts. We conducted the experiments based on the system model and our proposed framework introduced in Secs. ~\ref{sec:systemmodel} and \ref{sec: Proposed_Framework}. The duration of the initial phase is set to $L=10$, the incoming batch size $B_i = 64, \forall i \in \mathcal{N}$, $K_{\mathrm{min}} = 15$, softmax threshold $S_{\mathrm{th}}=0.25$ and significance level $\alpha=0.05$. At the \gls{es} side, the received samples are stored in a memory buffer that can hold up to 3000 samples. Once full, any incoming samples randomly replace existing ones from the majority class within the buffer.

In each training iteration, for the CIFAR-10 dataset, lightweight data augmentation (random cropping and horizontal flipping) is conducted to the samples from the training batch; for SVHN data augmentation is not applied. In our experiments, the parameters are updated 20 times per each training batch for both datasets. We also employ focal loss which addresses class imbalance, and focuses more on learning samples that are harder to classify~\cite{focal}. The focal loss parameters are set to $\gamma_{\mathrm{FL}} = 2$, along with $\alpha_{\mathrm{FL}} = 0.8$ and $1-\alpha_{\mathrm{FL}}$ for the anomalous and normal classes, respectively.

To evaluate OCLADS, we consider four baseline schemes that illustrate the impact of update frequency and timing on inference accuracy: an \textit{All-update, Random-update, Oracle OCLADS} and \textit{No-update method}. The first one updates the device-side model after every incoming batch, whereas the Random-update method performs the same number of updates as OCLADS, but schedules them at random batches. The update timing in this baseline scheme follows a discrete Gaussian distribution. As such, the updates are not performed in response to detected shifts. The third method represents an oracle OCLADS variant with perfect shift detection (no false alarms or missed detections). This baseline illustrates the effect of correct update timing while performing fewer updates than OCLADS. The No-update method is a naive baseline where the device-side model is initialized using the same fine-tuned parameters as the other methods, but no further updates are performed (i.e., the model remains static).

\subsection{Numerical Results} \label{sec:inference}

In our experiments, a total of  54 (73) distribution shifts occur out of 755 (997) batches for the CIFAR-10 (SVHN) dataset. OCLADS correctly detects 39 (51) shifts for CIFAR-10 (SVHN), and incurs 35 and 47 false alarms, respectively. Consequently, the device-side model is updated 74 times for CIFAR-10 and 98 times for SVHN. Overall, for both datasets the proposed framework reduces the number of model updates to below 10\% of those performed by the All-update scheme.

We compare the learning efficacy of OCLADS to the baseline schemes using the average online macro F1-score over time. This online metric for batch $i$ is defined as
\begin{equation}
    F1_o(i) = \frac{1}{i}\sum_{s=1}^iF1(Y_s,\hat{Y_s}),
\end{equation}
where $Y_s$ and $\hat{Y_s}$ are the ground-truth and predicted labels from batch $s$. This metric is preferred in our experiments over the online accuracy introduced in \cite{oclmetrics} due to the severe class imbalance in our \gls{ad} task.

\begin{figure}[t]
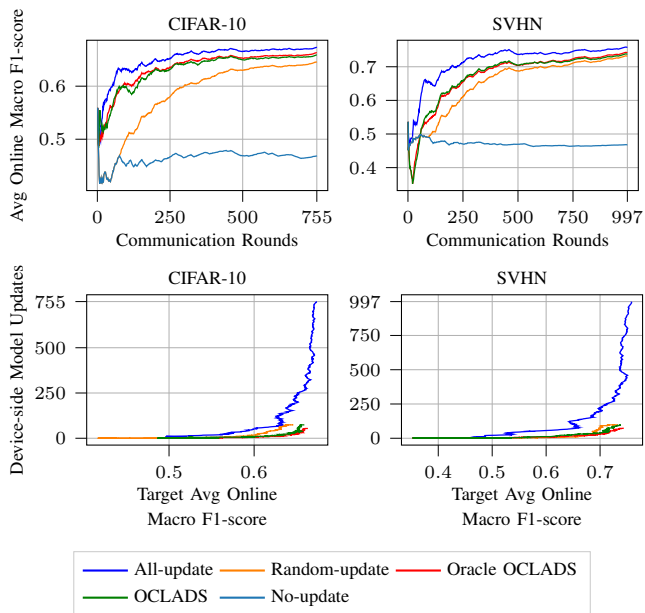

    \centering
    \begin{subfigure}[t]{\linewidth}
        \centering
        \begin{subfigure}[t]{0.47\linewidth}
            \input{figures_tex/binaryCifar/onlineF1_rounds}
        \end{subfigure}
        \hspace{1mm}
        \begin{subfigure}[t]{0.47\linewidth}
            \input{figures_tex/SVHN/onlineF1_rounds}
        \end{subfigure}
        \vspace{-4mm}
    \end{subfigure}
    \begin{subfigure}[t]{\linewidth}
        \centering
        \begin{subfigure}[t]{0.47\linewidth}
            \input{figures_tex/binaryCifar/onlineF1_updates}
        \end{subfigure}
        \hspace{1mm}
        \begin{subfigure}[t]{0.47\linewidth}
            \input{figures_tex/SVHN/onlineF1_updates}
        \end{subfigure}
        \vspace{-4mm}
        \begin{center}





\begin{tikzpicture}[inner sep=0pt, outer sep=0pt]

\definecolor{darkgray176}{RGB}{176,176,176}
\definecolor{darkturquoise0191191}{RGB}{0,191,191}
\definecolor{green01270}{RGB}{0,127,0}
\definecolor{lightgray204}{RGB}{204,204,204}
\definecolor{steelblue31119180}{RGB}{31,119,180}

\begin{axis}[
    hide axis,
    legend cell align={left},
    legend style={
        fill opacity=0.8,
        draw opacity=1,
        text opacity=1,
        draw=lightgray204
    },
    legend columns=3
]

\addplot[blue, thick] coordinates {(0,0)};
\addlegendentry{All-update}

\addplot[orange, thick] coordinates {(0,0)};
\addlegendentry{Random-update}

\addplot[red, thick] coordinates {(0,0)};
\addlegendentry{Oracle OCLADS}

\addplot[green01270, thick] coordinates {(0,0)};
\addlegendentry{OCLADS}

\addplot[steelblue31119180, thick] coordinates {(0,0)};
\addlegendentry{No-update}

\end{axis}
\end{tikzpicture}
        \end{center}
    \end{subfigure}
    \caption{Average online macro F1-scores w.r.t. the number of communication rounds and device-side model updates.}
    \label{onlineF1_figures}
\end{figure}

Fig. \ref{onlineF1_figures} illustrates the performance of OCLADS and the baseline methods with respect to the number of communication rounds and device-side model updates. For both datasets, the All-update scheme performs best, but also requires the largest number of device-side model updates. The No-update method achieves the lowest average online macro F1-scores, as the device-side model is insufficiently trained and not adapted to the dynamic conditions. The results also confirm that OCLADS achieves performance comparable to the All-update method with a significantly lower number of device-side model updates. Additionally, the performance of the Random-update scheme indicates that the timing is even more important than the number of model updates. It further validates the OCLADS framework by showing that updating the model after distribution shifts is more effective than performing updates at random. Moreover, the learning curve of the Oracle OCLADS suggests that with perfect shift detection, the performance of OCLADS would remain similar while updating the device-side model after only about 7\% of the batches.

Overall, these results verify that OCLADS provides a strong trade-off between communication efficiency and model performance in the most critical operating online \gls{cl} setting, i.e., when both distribution shifts and anomalies are present.

\balance

\section{Conclusion} \label{sec:conclusion}

This work introduced a novel TinyML-compatible framework for communication-efficient on-device inference with server-based online continual learning. By integrating a mechanism for selective data transmission and distribution shift detection, the device transmits only the most informative samples to the server for model updates, and obtains updated models only upon detected shifts. As such, the framework significantly lowers communication costs, while preserving robust performance in non-stationary settings.

Future work will focus on extending the framework to multi-device setup, and implementing image compression and retrieval methods to further improve communication efficiency. In addition, integrating power-saving communication techniques, such as wake-up radios or discontinuous reception, can be explored to further reduce device-side energy consumption.

\bibliographystyle{ieeetr}
\bibliography{bibliography}
\end{document}